\documentclass[sigconf]{acmart}

\AtBeginDocument{%
  \providecommand\BibTeX{{%
    \normalfont B\kern-0.5em{\scshape i\kern-0.25em b}\kern-0.8em\TeX}}}

\copyrightyear{2024}
\acmYear{2024}
\setcopyright{acmlicensed}
\acmConference[KDD '24] {Proceedings of the 30th ACM SIGKDD Conference on Knowledge Discovery and Data Mining }{August 25--29, 2024}{Barcelona, Spain.}
\acmBooktitle{Proceedings of the 30th ACM SIGKDD Conference on Knowledge Discovery and Data Mining (KDD '24), August 25--29, 2024, Barcelona, Spain}
\acmISBN{979-8-4007-0490-1/24/08}
\acmDOI{10.1145/3637528.3671945}

\usepackage{enumitem}
\usepackage{subcaption}
\usepackage{colortbl}
\usepackage{marvosym}

\def \algo {{Candle}}

\definecolor{darkgreen}{rgb}{0.0,0.7,0.0}
\def\red{\color{red}}
\def\green{\color{darkgreen}}

\usepackage{pifont}       
\newcommand{\cmark}{\ding{51}}
\newcommand{\xmark}{\ding{55}}

\begin{document}

\title{Efficient and Long-Tailed Generalization for Pre-trained Vision-Language Model}

\author{Jiang-Xin Shi}
\authornote{Equal contribution.}
\affiliation{%
  \department{National Key Laboratory for Novel Software Technology}
  \department{School of Artificial Intelligence}
  \institution{Nanjing University, China}
  \country{}
}
\email{shijx@lamda.nju.edu.cn}
\author{Chi Zhang}
\authornotemark[1]
\affiliation{%
  \department{National Key Laboratory for Novel Software Technology}
  \institution{Nanjing University, China}
  \country{}
}
\email{chi-zhang@smail.nju.edu.cn}
\author{Tong Wei}
\affiliation{%
  \department{School of Computer Science and Engineering}
  \department{Key Laboratory of Computer Network and Information Integration of Ministry of Education}
  \institution{Southeast University, China}
  \country{}
}
\email{weit@seu.edu.cn}
\author{Yu-Feng Li}
\authornote{Corresponding author.}
\affiliation{%
  \department{National Key Laboratory for Novel Software Technology}
  \department{School of Artificial Intelligence}
  \institution{Nanjing University, China}
  \country{}
}
\email{liyf@lamda.nju.edu.cn}

\renewcommand{\shortauthors}{Jiang-Xin Shi, Chi Zhang, Tong Wei, and Yu-Feng Li}

\begin{abstract}
Pre-trained vision-language models like CLIP have shown powerful zero-shot inference ability via image-text matching and prove to be strong few-shot learners in various downstream tasks. However, in real-world scenarios, adapting CLIP to downstream tasks may encounter the following challenges: 1) data may exhibit long-tailed data distributions and might not have abundant samples for all the classes; 2) There might be emerging tasks with new classes that contain no samples at all. To overcome them, we propose a novel framework to achieve efficient and long-tailed generalization, which can be termed as \textbf{\algo}. During the training process, we propose compensating logit-adjusted loss to encourage large margins of prototypes and alleviate imbalance both within the base classes and between the base and new classes. For efficient adaptation, we treat the CLIP model as a black box and leverage the extracted features to obtain visual and textual prototypes for prediction. To make full use of multi-modal information, we also propose cross-modal attention to enrich the features from both modalities. For effective generalization, we introduce virtual prototypes for new classes to make up for their lack of training images. \textbf{\algo}\ achieves state-of-the-art performance over extensive experiments on 11 diverse datasets while substantially reducing the training time, demonstrating the superiority of our approach. The source code is available at \url{https://github.com/shijxcs/Candle}.
\end{abstract}




\maketitle

\section{Introduction}
\label{sec:intro}
\quad 
Over the past few years, the rapid development of deep learning \cite{2017attention, dosovitskiy2021an} and the emergence of web-scale datasets \cite{fei2009imagenet, schuhmann2022laion, LAION400M} have made large-scale pre-trained models possible. Particularly, Vision-Language (V-L) models \cite{radford2021learning, jia2021scaling, li2022blip, brown2020language, yu2022coca} have become a recent research hype due to their strong generalization capabilities as well as promising transferability to downstream tasks. One of the most successful pre-trained V-L models is CLIP \cite{radford2021learning}. Trained on a massive dataset of 400 million image-text pairs, CLIP utilizes a contrastive objective to align the visual and textual representations and manages to establish a connection between images and natural language. During inference, CLIP can perform zero-shot image recognition by simply using the class names. For example, one can adopt a prompt template like `a photo of a \{class\}' as input to the text encoder and generate the classification weights for each class. The weights can then be used to calculate cosine similarity with image features to get classification scores.

With the rise of such powerful V-L models, extensive efforts have been invested into finding potential solutions to better adapt these models to downstream tasks. For instance, several previous works including CoOp\cite{zhou2022coop}, CoCoOp \cite{zhou2022cocoop} and MaPLe \cite{khattak2023maple} have explored the idea of prompt learning, where a sequence of learnable context vectors is used to replace carefully-chosen hard prompts. These methods have achieved impressive improvements. 

Despite delivering promising results, a number of existing works suffer from two practical limitations. a) significant performance decline under real-world long-tailed data distributions. The natural long-tail distribution \cite{openlongtailrecognition} phenomenon brings class imbalance and makes it hard to collect data for all classes, leaving some rare classes entirely void of samples. Most works fail to consider the real-world data distributions and suffer from severe performance degradation under imbalanced scenarios. They also tend to overlook the valuable label information for unseen categories, which may be a leading cause for a notable performance drop on these label-only classes. b) extensive computational overhead. Despite using fewer trainable parameters, most methods still need to calculate gradients through the model's backbone and require access to the model's weights. As the size of foundation V-L models continues to grow (\textit{e.g.}, up to 80 billion \cite{brown2020language}) and industry standards gradually switch to providing only API, they may become impractical for actual application. 

\begin{figure}[!t]
\begin{minipage}[t]{\textwidth}
    \includegraphics[width=0.49\linewidth]{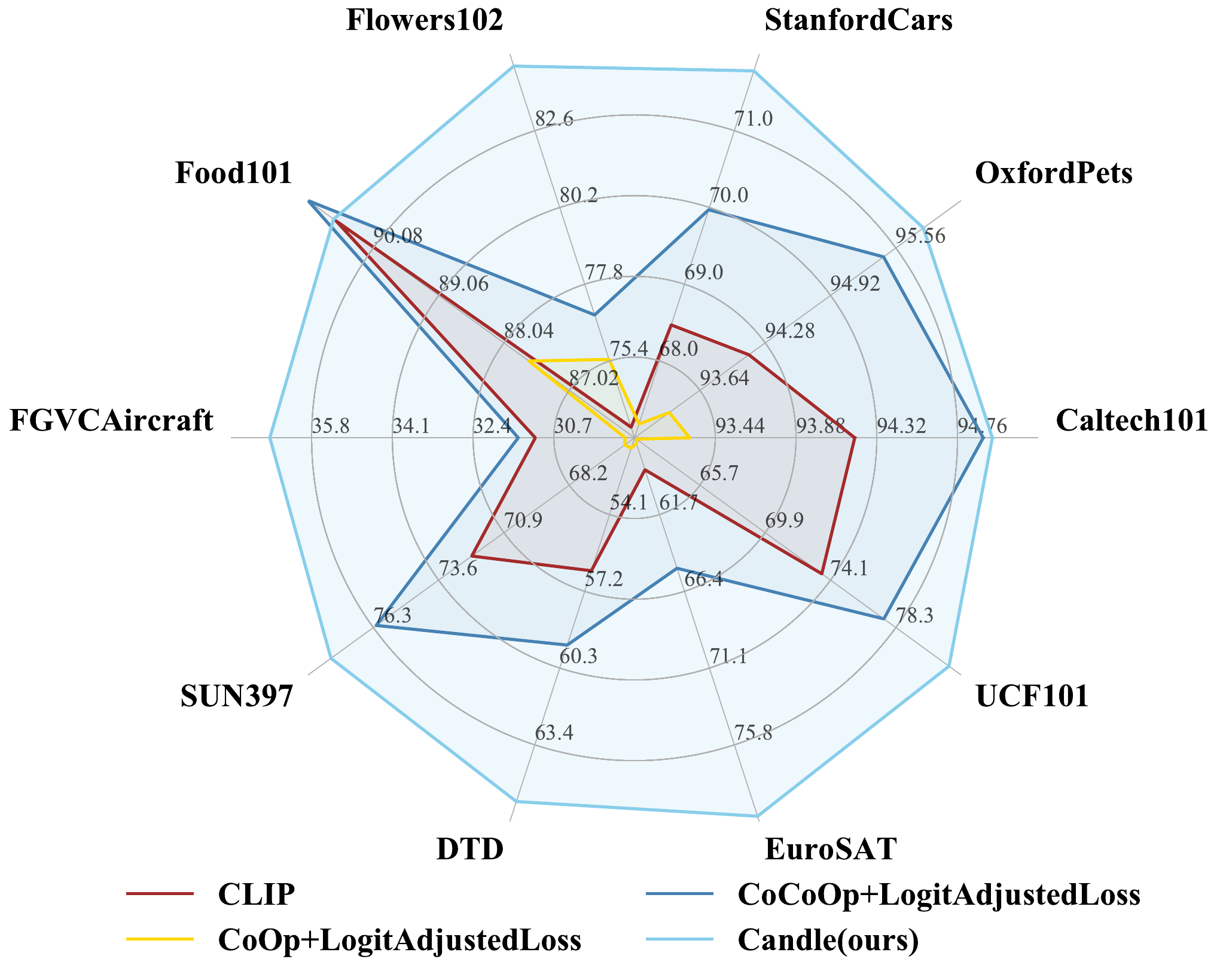}
\end{minipage}
\vskip -0.05in
\caption{\algo \ achieves significant improvements on multiple imbalanced base-to-new generalization tasks.}
\label{fig:teaser}
\end{figure}

\begin{table}[!t]
\caption{Training time and accuracy for CoOp, CoCoOp, and \algo\, with ViT-B/16 as the visual encoder, on a 100-shot ImageNet with an imbalance ratio of 100. The experiments are done on an RTX 3090 GPU.}
\vskip -0.05in
\label{tab:teaser}
\begin{center}
\begin{tabular}{c|ccc}
\toprule
Method & Training epochs & Time cost & Accuracy (\%) \\ \midrule
CoOp & 50 & $\sim$ 5 hours & 70.7 \\
CoCoOp & 10 & $\sim$ 30 hours  & 71.3 \\
\midrule
\rowcolor[HTML]{E8E5E5} \algo~(ours) & 20 & \textbf{11 min} & \textbf{71.6} \\
\bottomrule
\end{tabular}
\end{center}
\end{table}

In this paper, we aim to address the above issues and propose a novel framework to achieve efficient and long-tailed generalization which can be named as \algo. During the training process, we propose compensating logit-adjusted loss to encourage large margins of virtual prototypes and alleviate imbalance both within the base classes and between the base and new classes. For efficient adaptation, we treat the CLIP model as a black box and leverage the extracted features to obtain visual and textual prototypes for prediction. To make full use of multi-modal information, we also propose cross-modal attention to enrich the features from both modalities. For effective generalization, we introduce virtual prototypes for new classes to make up for their lack of training images. As shown in Figure~\ref{fig:teaser} and Table~\ref{tab:teaser}, our method achieves impressive improvements over previous methods while cutting down the training time. In summary, the main contributions of this work include:

\begin{itemize}[leftmargin=2ex, itemindent=3ex, topsep=1.5ex]
    \item We propose a novel framework named \algo\ for efficient and long-tailed generalization of CLIP. To the best of our knowledge, this is the first work to explore the adaptation of V-L models under an imbalanced setting.
    \item To make full use of both visual and textual information, we propose to perform \textit{cross-modal attention} on the feature space. For better new class generalization, we introduce \textit{virtual prototypes} and propose a novel \textit{compensating logit adjusted loss} to simultaneously alleviate the imbalance within the base classes as well as between the base and new classes.
    \item Our extensive experimental results demonstrate the strength of \algo, which achieves state-of-the-art results over various settings while substantially reducing the training time.
\end{itemize}

\section{Related Work}

\textbf{Vision-Lanuage (V-L) Models.}
V-L foundation models have experienced a substantial surge in recent years with the emergence of different architectures such as Flamingo \cite{brown2020language}, CLIP \cite{radford2021learning}, ALIGN \cite{jia2021scaling},  BLIP \cite{li2022blip}, CoCa \cite{yu2022coca}, etc.
These models are usually trained on a web-scale dataset comprised of massive image-text pairs to learn a joint embedding space. Due to their strength in understanding open-vocabulary concepts, V-L models have been widely explored in various downstream tasks, such as few-shot learning \cite{song2022clip, zhou2022coop}, continual learning \cite{ding2022don, zhou2023learning} and adversarial learning \cite{tao2023galip, mao2022understanding}. In this work, we focus on adapting CLIP for new class generalization. 

\textbf{Fine-tuning V-L Models.} 
Despite the effectiveness of V-L models (\textit{e.g.}, CLIP) towards generalizing to new concepts, its massive scale makes it infeasible to fine-tune the full model for downstream tasks. Linear probing \cite{radford2021learning} serves as a naive solution, while its performance deteriorates significantly under few-shot settings. CoOp \cite{zhou2022coop} proposes the idea of prompt learning, which optimizes a set of context vectors instead of using the standard prompt template `a photo of a \{class\}'. CoCoOp \cite{zhou2022cocoop} aims to learn more robust prompts through image conditioning, which optimizes an instance-specific prompt by training a meta-network. CoCoOp also proposes the novel base-to-new setting for better examination of a model's generalizability. MaPLe \cite{khattak2023maple} simultaneously learns the prompts for both the vision and language branches of CLIP. While these methods have achieved impressive results under few-shot settings, their training cost can be prohibitive in terms of both time and memory.

Aside from prompt learning, another line of work utilizes adapter modules for lightweight and fast adaptation. For instance, CLIP-Adapter \cite{gao2021clip} proposes to add an MLP layer after the final visual layer and mix the transformed output with the original zero-shot output via a residual connection. TIP-Adapter \cite{zhang2021tip} further replaces the MLP layer with a carefully designed linear layer, whose weights are comprised of labeled visual embeddings. Although these works have significantly reduced the training cost for fine-tuning CLIP, they perform poorly under the base-to-new setting, with TIP-Adapter \cite{zhang2021tip} even unable to test on new classes.

\begin{table*}[!t]
\caption{Empirical study results for zero-shot CLIP and visual prototypes over 11 datasets, using ViT-B/16 as the visual encoder. The visual prototypes are obtained by calculating the mean value of 16-shot features for each class and used subsequently to calculate cosine similarity with image features to get the classification scores.}
\vskip -0.05in
\label{tab: vp}
\begin{tabular}{@{}ccccccccccccc@{}}
\toprule
& CAL.          & OP.           & SC.           & Flw.          & Food.         & FA.           & SUN.          & DTD.          & ES.           & UCF.          & IN.           & Avg Results.          \\ \midrule
\multicolumn{1}{c|}{Zero-shot CLIP}    & 89.3          & \textbf{88.9} & 65.6          & 70.4          & \textbf{89.2} & 27.1          & 65.2          & 46.0          & 54.1          & 69.8          & \textbf{68.6} & 66.7          \\
\multicolumn{1}{c|}{Visual Prototypes} & \textbf{93.4} & 80.2          & \textbf{71.7} & \textbf{95.9} & 81.4          & \textbf{41.3} & \textbf{69.8} & \textbf{64.2} & \textbf{75.2} & \textbf{78.1} & 61.7          & \textbf{73.9} \\ 
\multicolumn{1}{c|}{$\Delta$} &\green +4.1 & \red -8.7 & \green +6.1 & \green +25.5 & \red -7.8 & \green +14.2 & \green +4.6 & \green +18.2 & \green +21.1 & \green +8.3 & \red -6.9 & \green +7.2\\
\bottomrule
\end{tabular}
\end{table*} 

Furthermore, subsequent works have attempted to improve adaptation by leveraging multi-modal information \cite{p2023protoclip} or adopting a generative approach to synthesize features for categories without data \cite{wang2023improving}. However, they still suffer from expensive training costs or unsatisfactory new-class generalizations. In contrast to the aforementioned works, this paper presents a lightweight framework built directly upon the feature space for efficient adaptation, as well as virtual prototypes with a novel loss function for effective new class generalization. 

\textbf{Imbalance Learning via Pre-trained Models.} 
Recent research has found that models pre-trained on large-scale datasets can learn more generalized representations, and can serve as an effective tool for alleviating class imbalance issues. For instance, BALLAD \cite{ma2021simple} and VL-LTR \cite{tian2022vl} fine-tunes both of the entire image and text encoder of CLIP on the downstream tasks. \citet{wang2023exploring} proposes to ignore the text branch of CLIP and append a decoder consisting of transformer blocks after the image encoder. LPT \cite{dong2023lpt} adopts a two-stage method to learn both shared prompts and group-specific prompts to capture both general and specialized knowledge. PEL \cite{shi2023parameter} systematically investigates different parameter-efficient fine-tuning modules for long-tailed recognition tasks. 
In contrast to these previous works, this paper deals with the new class generalization setting and proposes an efficient and effective approach.
\label{sec:related}

\section{CLIP}
\label{sec:revisiting CLIP}

\subsection{Premilinaries}
\label{sec: zsclip}
\quad Contrastive Language-Image Pretraining, known as CLIP \cite{radford2021learning}, is mainly comprised of an image encoder $f_{I}(x)$ and a text encoder $f_{T}(t)$, which map input from the respective modality into a joint embedding space. The image encoder can be in the form of either ResNet \cite{he2016deep} or ViT \cite{dosovitskiy2021an}, whereas the text encoder is built on top of the Transformer \cite{2017attention} architecture. 

During training, CLIP goes through 400 million image and caption pairs, adopting a contrastive loss to pull together the corresponding image-text pairs while pushing apart unmatched ones. After training, CLIP can be readily used for downstream image classification in a zero-shot manner. Let $x$ be the input image and $\{t_{1}, \cdots, t_{K}\}$ be the $K$ class descriptions. These descriptions can be generated through prompt templates like `a photo of a \{class\}', where the \{class\} token denotes the corresponding class name. Then, it extracts image features $\boldsymbol{x}=f_{I}(x)$, textual prototypes $\boldsymbol{T}=\{\boldsymbol{T}_{1}, \cdots, \boldsymbol{T}_{K}\}=\{f_T(t_{1}), \cdots, f_T(t_{K})\}$, and the predicted result for $x$ is:
\begin{equation}
  y_{\mathrm{pred}} = \arg\max_{i\in[K]}\cos(\boldsymbol{x}, \boldsymbol{T}_{i})
  \label{eq:zsclip}
\end{equation}
In this way, CLIP turns the image classification task into an image-text matching problem.

\subsection{Practical Limitations of CLIP}
Despite delivering promising results, a number of existing works suffer from two practical limitations. a) significant performance decline under real-world long-tailed data distributions. The natural long-tail distribution \cite{openlongtailrecognition} phenomenon brings class imbalance and makes it hard to collect data for all classes, leaving some rare classes entirely void of samples. Most works fail to consider the real-world data distributions and suffer from severe performance degradation under imbalanced scenarios. They also tend to overlook the valuable label information for unseen categories, which may be a leading cause for a notable performance drop on these label-only classes. b) extensive computational overhead. Despite using fewer trainable parameters, most methods still need to calculate gradients through the model's backbone and require access to the model's weights. As the size of foundation V-L models continues to grow (\textit{e.g.}, up to 80 billion \cite{brown2020language}) and industry standards gradually switch to providing only API, they may become impractical for actual application. 

Moreover, despite the effectiveness of CLIP in general cases, its insufficient usage of visual information remains a weak point. CLIP relies highly on image-text matching for downstream zero-shot prediction, which may cause potential risks. For instance, on the FGVCAircraft \cite{maji2013fgvc} dataset, the class names are different numeral versions such as `737-200' and `737-300', which hardly contain any useful information; or on the UCF101 \cite{soomro2012ucf101} dataset, the image samples consist of frames from a video and do not precisely match the prompt templates such as `a photo of a \{class\}'. 

Based on this motivation, we conduct our empirical study by comparing image-image matching with image-text matching. Specifically, we calculate visual prototypes as the mean value of the 16-shot image features for each class. Then, we replace the textual prototypes in zero-shot CLIP with visual prototypes for prediction. Formally, let $V=\{\boldsymbol{V}_{1}, \cdots, \boldsymbol{V}_{K}\}$ be the $K$ visual prototypes for each class,  then the predicted result is:
\begin{equation}
  y_{\mathrm{pred}} = \arg\max_{i\in[K]}\cos(\boldsymbol{x}, \boldsymbol{V}_{i})
  \label{eq:vp}
\end{equation} 
\quad We compare the prediction results calculated by Equation~\ref{eq:zsclip} and by Equation~\ref{eq:vp}, and report the empirical results in Table~\ref{tab: vp}. As shown, zero-shot CLIP significantly underperforms visual prototypes on multiple datasets including Flowers102 (-25.5\%), FGVCAirCraft (-14.2\%), DTD (-18.2\%) and EuroSAT (-21.1\%), demonstrating that the class labels are not descriptive enough.

\begin{figure*}[!t]
    \centering
    \includegraphics[width=0.95\textwidth]{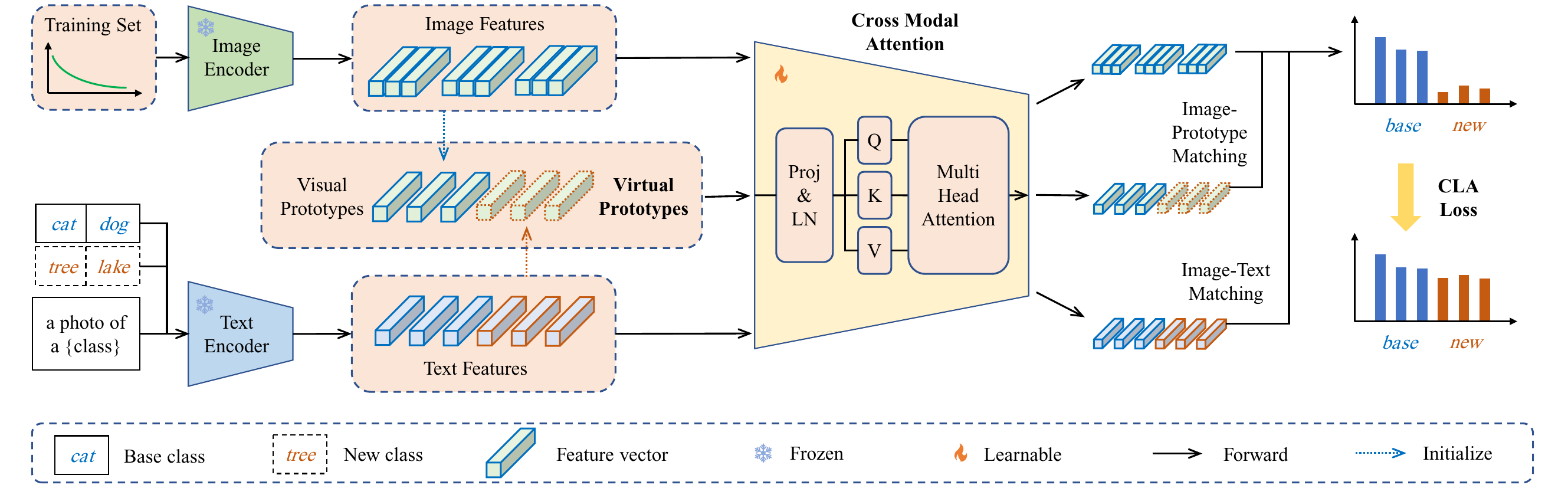}
    \vskip -0.05in
    \caption{An overview of the proposed framework.}
    \label{fig:framework}
\end{figure*}

\section{\algo: Efficient and Long-Tailed Generalization}
\label{sec:method}
In this section, we aim to address the above issues and propose a novel framework to achieve efficient and long-tailed generalization which can be named as \algo. During the training process, we propose compensating logit-adjusted loss to mitigate the long-tail problems, as well as to avoid the risk of neglecting new classes during the optimization. For efficient adaptation, we directly leverage the features extracted from the model, calculate the corresponding visual and textual prototypes, and propose cross-modal attention to enrich the information for both modalities. For effective generalization, we propose to generate virtual prototypes for new classes, by which we compensate for their lack of training images.

\subsection{Compensating Logit-Adjusted Loss}
\quad Loss functions designed to deal with class imbalance usually do not apply to new class generalization since there is no sample for new classes. Therefore, we propose to consider new class generalization as an extreme case of class imbalance and treat visual prototypes as samples to alleviate such imbalance. 

On top of this, we introduce our Compensating Logit-Adjusted Loss (CLA Loss) inspired by \cite{ren2020balanced} to handle such imbalanced scenarios. Let $z_{i}$ denote the predicted logit for class $i$, then CLA loss takes the form of :
\begin{equation}
\mathcal{L}_{cla}(\boldsymbol{z}, y = j) = -\log \frac{\exp(z_{j} + \log p(y=j))}{\sum_{k=1}^{K} \exp(z_{k} + \log p(y=k))}
  \label{eq:cla}   
\end{equation}
where $p(y=j)$ is the class prior probability. Let $n_{i}$ denote the number of samples for class $i$, $p(y=j)$ can be estimated as $n_{j} / \sum_{i=1}^{K} n_{i}$. Particularly, we suppose $n_{i} = 1$ for new classes and treat the corresponding visual prototypes as training samples. In this way, we have found a solution to not only deal with the imbalance within the base classes but also compensate for the imbalance between the base and new classes.

\subsection{Feature-Level Cross-Modal Attention}
\label{sec: cma}

\quad To overcome the efficiency issue, we introduce cross-modal attention to leverage both visual and textual information. To further enhance its efficiency and practicality, we treat the model as a black box following \cite{Ouali2023blackbox} and apply optimizations directly within the feature space. 

Our method can divided into the following steps. First, we pre-compute and save the visual and textual prototypes for each class. Then, the features together with the prototypes are fed into the corresponding linear projection layers $P_I$ and $P_T$. In this way, we can further align the two modalities for downstream adaptation, and the prediction can be calculated by
\begin{equation}
  p(y = i \mid x) = \frac{\exp(\cos(P_{I}(\boldsymbol{x}), P_{T}(\boldsymbol{T}_{i})) / \tau_{t})} {\sum_{j=1}^{N} \exp(\cos(P_{I}(\boldsymbol{x}), P_{T}(\boldsymbol{T}_{j})) / \tau_{t})}
  \label{eq:lp}
\end{equation}
where $\tau_{t}$ is the temperature for image-text matching. However, the scarce data in the downstream tasks still makes it difficult to achieve satisfactory adaptations.

To remedy this issue, we propose to enrich the features from both modalities by leading them to interact with each other through cross-modal attention. Specifically, we concatenate the image features, visual prototypes, and textual prototypes together and then feed them into a self-attention \cite{2017attention} module, considering its ability to establish connections for long-dependency embeddings. Let $\operatorname{Attn(\cdot)}$ denote the multi-head self-attention function, the output is
\begin{equation}
    \boldsymbol{x}', \boldsymbol{V}', \boldsymbol{T}' = \operatorname{Attn}([P_{I}(\boldsymbol{x}), P_{I}(\boldsymbol{V}), P_{T}(\boldsymbol{T})])
\end{equation}
Note that the operation is permutation invariant, thereby we can slice the output to get the corresponding features and prototypes. After obtaining the enriched features, we can now predict with both visual and textual prototypes by:
\begin{equation}
  p_{V}(y = i \mid x) = \frac{\exp(\cos(\boldsymbol{x}', \boldsymbol{V}'_{i}) / \tau_{v})} {\sum_{j=1}^{K} \exp(\cos(\boldsymbol{x}', \boldsymbol{V}'_{j}) / \tau_{v})} 
  \label{eq:attn-v}
\end{equation}
\begin{equation}
  p_{T}(y = i \mid x) = \frac{\exp(\cos(\boldsymbol{x}', \boldsymbol{T}'_{i}) / \tau_{t})} {\sum_{j=1}^{K} \exp(\cos(\boldsymbol{x}', \boldsymbol{T}'_{j}) / \tau_{t})} 
  \label{eq:attn-t}
\end{equation}
where $\tau_{v}$ denotes the temperature for visual modality. By ensembling these two results, we are able to leverage both visual and textual information.

\subsection{Virtual Prototypes for New Classes}
\label{sec:vp}
\quad We have discussed the weakness of CLIP and introduced visual prototypes as well as cross-modal attention as a remedial measure. However, another crucial problem emerges regarding new class generalization. As for new classes, we cannot obtain their visual prototypes during training. 

To solve this issue, we introduce learnable virtual prototypes for new classes to hold the place of missing visual prototypes. Specifically, we freeze the precomputed textual prototypes as well as the visual prototypes for base classes, while treating the virtual prototypes as the corresponding visual prototypes for new classes, and optimizing them during the training stage. Other than this, the entire procedure is the same as in Section~\ref{sec: cma}.  Formally, let $\hat{\boldsymbol{V}}'$ denote the transformed virtual prototypes for new classes, then Equation~\ref{eq:attn-v} can be rewritten as:
\begin{equation}
p(y = i \mid w) = \frac{\exp(\cos(\boldsymbol{x}', \boldsymbol{V}'_{i}) / \tau_{v})} {
\sum \exp (\frac{\cos(\boldsymbol{x}', \boldsymbol{V}')}{\tau_{v}}) + \sum \exp(\frac{\cos(\boldsymbol{x}', \hat{\boldsymbol{V}}')}{ \tau_{v}})} 
  \label{eq:attn-b2n}   
\end{equation}
In this way, the virtual prototypes are guided to learn a suitable representation of the new classes in the feature space and can act as an supplement to the textual prototypes. 

\subsection{Overall Objective}
The overall loss function is given by applying CLA loss to the logits calculated by Equation~\ref{eq:lp},~\ref{eq:attn-v} and~\ref{eq:attn-t} and aggregating the results. Suppose the logits given by these equations are $\boldsymbol{z}_{P}, \boldsymbol{z}_{V}, \boldsymbol{z}_{T}$ respectively, then the loss objective $\mathcal{L}$ for optimization during training is:
\begin{equation}
\mathcal{L} =  \mathcal{L}_{cla}(\boldsymbol{z}_{P}, y) +  \mathcal{L}_{cla}(\boldsymbol{z}_{V}, y) +   \mathcal{L}_{cla}(\boldsymbol{z}_{T}, y)
    \label{eq: loss}
\end{equation}
During inference, we aggregate the logits obtained after cross-modal attention, namely $\boldsymbol{z} = \boldsymbol{z}_{V} + \boldsymbol{z}_{T}$. The entire framework is shown in Figure~\ref{fig:framework}.

\section{Experiments} 

\subsection{Experimental Settings}

\quad
We evaluate our approach \algo\ in the following problem settings: 1) generalization from base to new classes under imbalance and few-shot settings; 2) cross-dataset transfer and 3) domain generalization. For the imbalanced settings, all the training data is generated by down-sampling the base classes to obey an exponential decay of different ratios. Let $n_{i}$ denote the number of samples in the $i$-th class, imbalance ratio is defined as $\max\{n_{i}\} / \min\{n_{i}\}$. The maximum number of samples per class of the generated dataset is set to either 100 (if has) or the maximum number of samples per class of the original dataset.

\textbf{Datasets and Evaluation.}
For new class generalization and cross-dataset transfer, the experiments are conducted over a total of 11 diverse image classification datasets, including ImageNet \cite{fei2009imagenet} and Caltech101 \cite{fei2004caltech} for generic object recognition, OxfordPets \cite{2012pets}, StanfordCars \cite{fei2013cars}, Flowers102 \cite{2008flowers}, Food101 \cite{bossard2014food} and FGVCAircraft \cite{maji2013fgvc} for fine-grained image recognition, SUN397 \cite{2010SUN} for scene recognition, DTD \cite{2014DTD} for texture classification, EuroSAT \cite{2019eurosat} for satellite image classification, and UCF101 \cite{soomro2012ucf101} for action recognition. For domain generalization, we use ImageNet as the source dataset and four other variants that exhibit different types of domain shift as the target datasets, including ImageNet-A \cite{hendrycks2021nae}, ImageNetV2 \cite{recht2019imagenet}, ImageNet-Sketch \cite{wang2019learning}, and ImageNet-R \cite{hendrycks2021many}.

Details of the 11 datasets used in base-to-new generalization and cross-dataset transfer, and the 4 datasets used during testing for domain generalization, are shown respectively in Table~\ref{tab:data1} and Table~\ref{tab:data2}. We report mean-class accuracy for the imbalanced settings, which is different from overall accuracy for datasets with imbalanced test sets. The test set for some datasets have varying numbers of sample per class, which is indicated in the rightmost column.
\begin{table}[!t]
\caption{Statistics for datasets used in base-to-new generalization and cross-dataset transfer. The rightmost column indicates whether the testing set is balanced.  }
\vskip -0.05in
\label{tab:data1}
\begin{center}
\begin{tabular}{c|cccc}
\toprule
Dataset      & Classes & Train & Test & Balanced \\ 
\midrule
ImageNet \cite{fei2009imagenet}     &   1000       &  1.28M     &  50000    &  \cmark      \\
Caltech101 \cite{fei2004caltech}   &   100      &  4128     &  2456    &  \xmark     \\
OxfordPets \cite{2012pets}  &   37      &  2944     &  3669    &  \cmark      \\
StanfordCars \cite{fei2013cars} &   196      &  6509     &  8041    &  \cmark \\
Flowers102  \cite{2008flowers} &   102      &  4093     &  2463    &  \xmark      \\
Food101 \cite{bossard2014food}     &   101      &  50500     &  30300    & \cmark       \\
FGVCAircraft \cite{maji2013fgvc} &   100      &  3334     &  3333    &   \cmark    \\
SUN397  \cite{2010SUN}     &   397      &  15880     & 19850     &  \cmark \\
DTD   \cite{2014DTD}       &   47      &  2820     &  1692    &  \cmark      \\
EuroSAT \cite{2019eurosat}     &   10      &  13500     &  8100    &  \xmark      \\
UCF101 \cite{soomro2012ucf101}     &   101      &  7639     &  3783    &  \xmark        \\ 
\bottomrule
\end{tabular}
\end{center}
\end{table}

\begin{table}[!t]
\caption{Statistics for datasets used during testing for domain generalization. The rightmost column indicates whether the testing set is balanced.  }
\vskip -0.05in
\label{tab:data2}
\tabcolsep=2ex
\begin{center}
\begin{tabular}{c|ccc}
\toprule
Dataset      & Classes  & Test & Balanced \\ 
\midrule
ImageNet-A \cite{hendrycks2021nae} & 200 & 7500 & \xmark \\
ImageNetV2 \cite{recht2019imagenet} & 1000 & 10000 & \cmark \\
ImageNet-Sketch \cite{wang2019learning} & 1000 & 50889 &  \cmark \\
ImageNet-R \cite{hendrycks2021many} & 200 & 30000 & \xmark \\
\bottomrule
\end{tabular}
\end{center}
\end{table}

Following the setting in CoCoOp \cite{zhou2022cocoop}, we examine our model on a similar but more practical scenario, where the base training set follows an imbalanced distribution. We also report test results of new class generalization in the balanced few-shot form to show the robustness of our model. Note that for imbalanced scenarios, we report mean-class accuracy instead of overall accuracy.

\textbf{Baselines.}
We compare our method to zero-shot CLIP \cite{radford2021learning}, CoOp \cite{zhou2022coop}, CoCoOp \cite{zhou2022cocoop} and LFA \cite{Ouali2023blackbox}, which also focuses on feature-level adaptation for CLIP. For the imbalanced settings, our method is compared to CoOp and CoCoOp by switching their loss function to Logit-Adjusted (LA) Loss \cite{ren2020balanced} to ensure fairness. LFA is only compared under the balanced setting because its framework is not compatible to different loss functions. 

\begin{table*}[!t]
\captionsetup{skip=0pt}
\caption{Harmonic mean values of base-to-new accuracy (\%) of different methods on datasets with imbalance ratios 10, 20, 50. The models are trained on an imbalanced base set and then evaluated on both base and new classes. Harmonic accuracy is calculated by $\frac{2 \times base \times new}{base + new}$ to highlight the generalization trade-off. The best results are presented in bold.}
\vskip -0.05in
\label{tab:ib2n}
\begin{subtable}{\textwidth}
\caption{\textbf{Imbalance Ratio = 10}.}
\begin{center}
\begin{tabular}{c|ccccccccccc} 
\toprule
& Cal. & OP. & SC. & FLw. & Food. & FA. & SUN. & DTD. & ES. & UCF. & Avg. \\
\midrule
CoOp + LogitAdjusted Loss & 91.78 & 94.10 & 69.23 & 71.86 & 89.25 & 32.12 & 72.15 & 54.88 & 54.42 & 64.74 & 70.66 \\
CoCoOp + LogitAdjusted Loss & 95.09 & \textbf{96.69} & 71.91 & 77.61 & \textbf{91.20} & 33.71 & 77.99 & 65.11 & 60.28 & 76.78 & 75.67 \\
Linear Feature Alignment & 95.69 & 94.09 & 72.72 & 84.38 & 90.44 & 34.27 & 78.39 & 67.43 & 69.56 & 82.71 & 76.97 \\
\rowcolor[HTML]{E8E5E5} \algo~(Ours) & \textbf{95.89} & 95.99 & \textbf{74.30} & \textbf{85.03} & 90.80 & \textbf{37.78} & \textbf{79.26} & \textbf{68.13} & \textbf{80.51} & \textbf{83.17} & \textbf{79.34} \\
\bottomrule
\end{tabular}
\end{center}
\end{subtable}
\begin{subtable}{\textwidth}
\caption{\textbf{Imbalance Ratio = 20}.}
\begin{center}
\begin{tabular}{c|ccccccccccc} 
\toprule
& Cal. & OP. & SC. & FLw. & Food. & FA. & SUN. & DTD. & ES. & UCF. & Avg. \\
\midrule
CoOp + LogitAdjusted Loss & 92.65 & 94.15 & 67.39 & 73.72 & 86.38 & 29.33 & 68.93 & 55.18 & 62.64 & 60.12 & 70.08 \\
CoCoOp + LogitAdjusted Loss & 95.25 & \textbf{96.64} & 71.38 & 80.31 & \textbf{91.20} & 32.78 & 77.29 & 61.31 & 58.82 & 71.70 & 74.48 \\
Linear Feature Alignment & 95.56 & 90.90 & 70.35 & 84.03 & 89.72 & 33.02 & 77.30 & 66.07 & 68.74 & 81.80 & 75.75 \\
\rowcolor[HTML]{E8E5E5} \algo~(Ours) & \textbf{95.84} & 95.89 & \textbf{73.49} & \textbf{84.92} & 90.75 & \textbf{38.02} & \textbf{78.53} & \textbf{67.32} & \textbf{80.96} & \textbf{82.59} & \textbf{79.08} \\
\bottomrule
\end{tabular}
\end{center}
\end{subtable}
\begin{subtable}{\textwidth}
\caption{\textbf{Imbalance Ratio = 50}.}
\begin{center}
\begin{tabular}{c|ccccccccccc} 
\toprule
& Cal. & OP. & SC. & FLw. & Food. & FA. & SUN. & DTD. & ES. & UCF. & Avg. \\
\midrule
CoOp + LogitAdjusted Loss & 93.30 & 93.34 & 67.18 & 75.45 & 87.65 & 29.20 & 65.91 & 51.42 & 57.35 & 61.62 & 69.92 \\
CoCoOp + LogitAdjusted Loss & 94.90 & 95.44 & 69.97 & 76.84 & \textbf{91.10} & 31.45 & 76.18 & 59.37 & 64.99 & 77.53 & 74.32 \\
Linear Feature Alignment & 94.23 & 86.76 & 67.95 & 82.81 & 87.73 & 30.75 & 75.13 & 61.78 & 61.91 & 79.49 & 72.85 \\
\rowcolor[HTML]{E8E5E5} \algo~(Ours) & \textbf{94.95} & \textbf{95.83} & \textbf{71.78} & \textbf{84.62} & 90.70 & \textbf{36.68} & \textbf{78.05} & \textbf{65.69} &  \textbf{80.17} & \textbf{81.72} & \textbf{78.20} \\
\bottomrule
\end{tabular}
\end{center}
\end{subtable}
\end{table*}

\begin{figure*}[!t]
    \centering
    \includegraphics[width=0.31\textwidth]{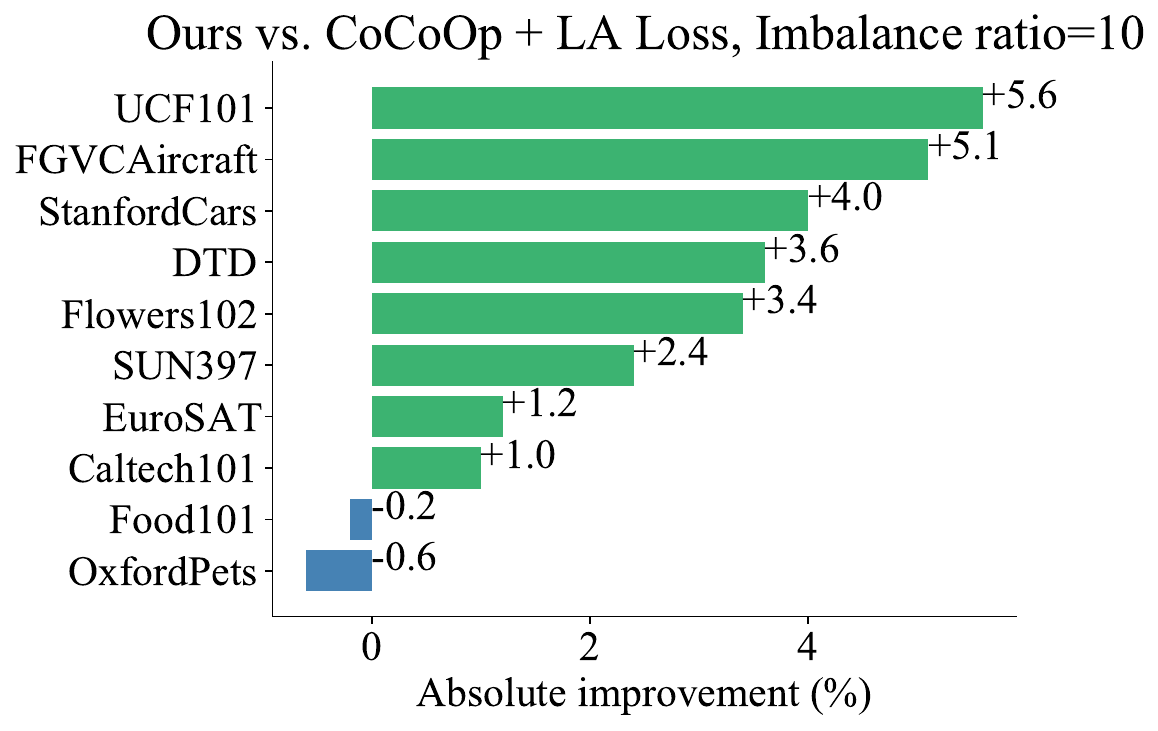}
    \includegraphics[width=0.31\textwidth]{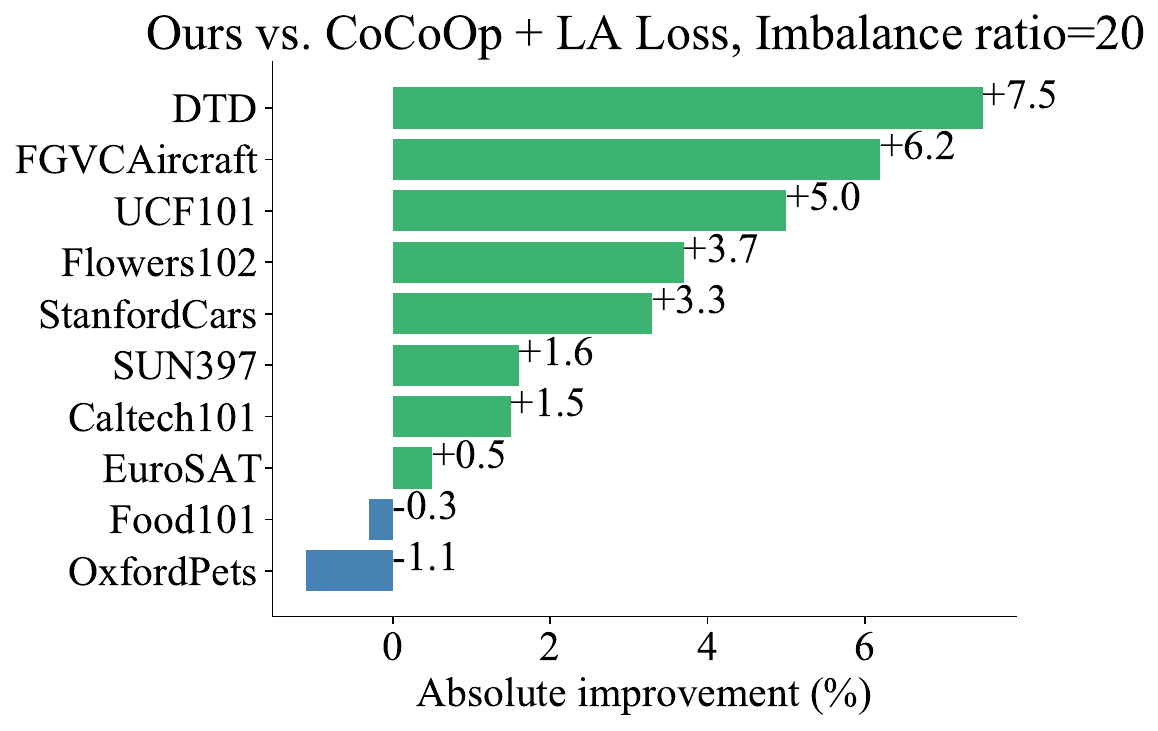}
    \includegraphics[width=0.31\textwidth]{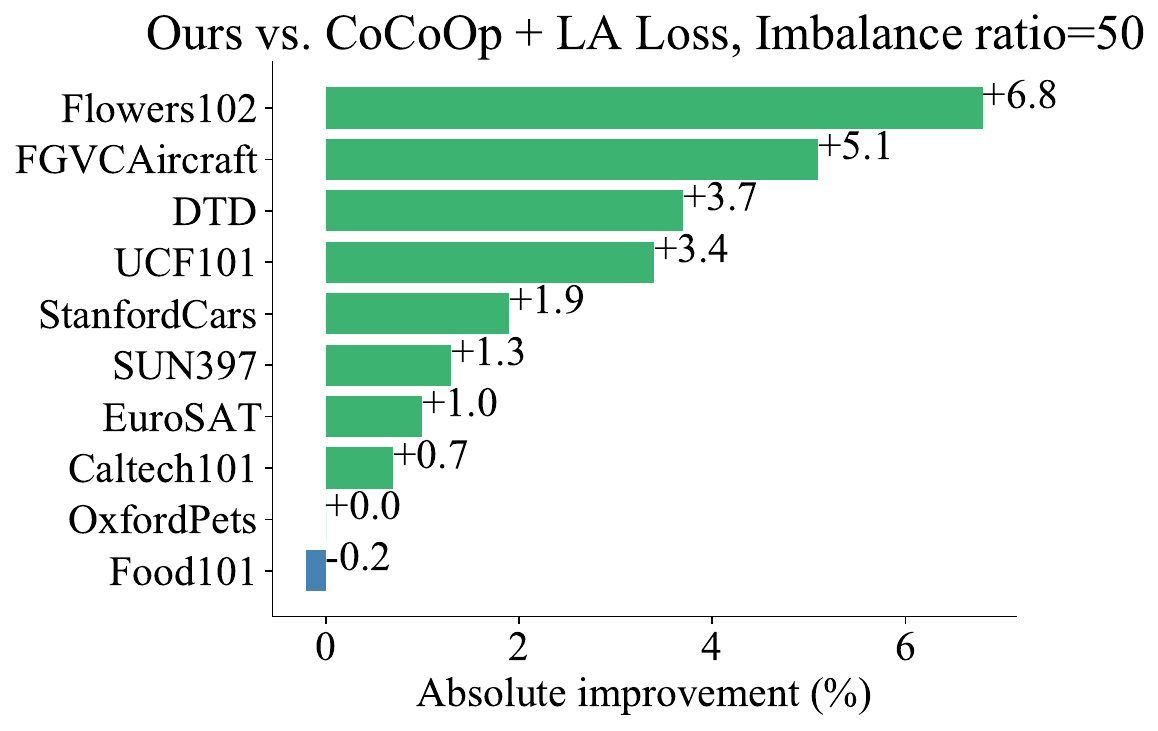}
    \vskip -0.1in
    \caption{Absolute improvement on the base classes with imbalance ratio 10, 20, 50}
    \label{fig:ib2n1}
\end{figure*}

\begin{figure*}[!t]
    \centering
    \includegraphics[width=0.31\textwidth]{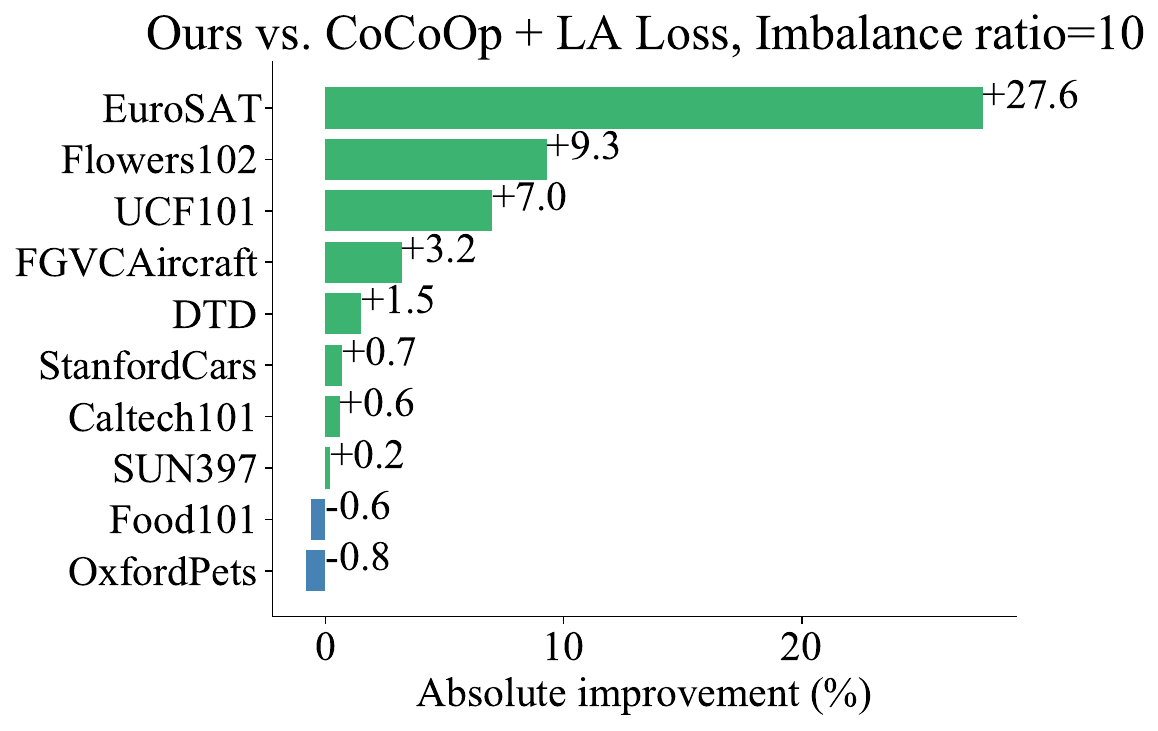}
    \includegraphics[width=0.31\textwidth]{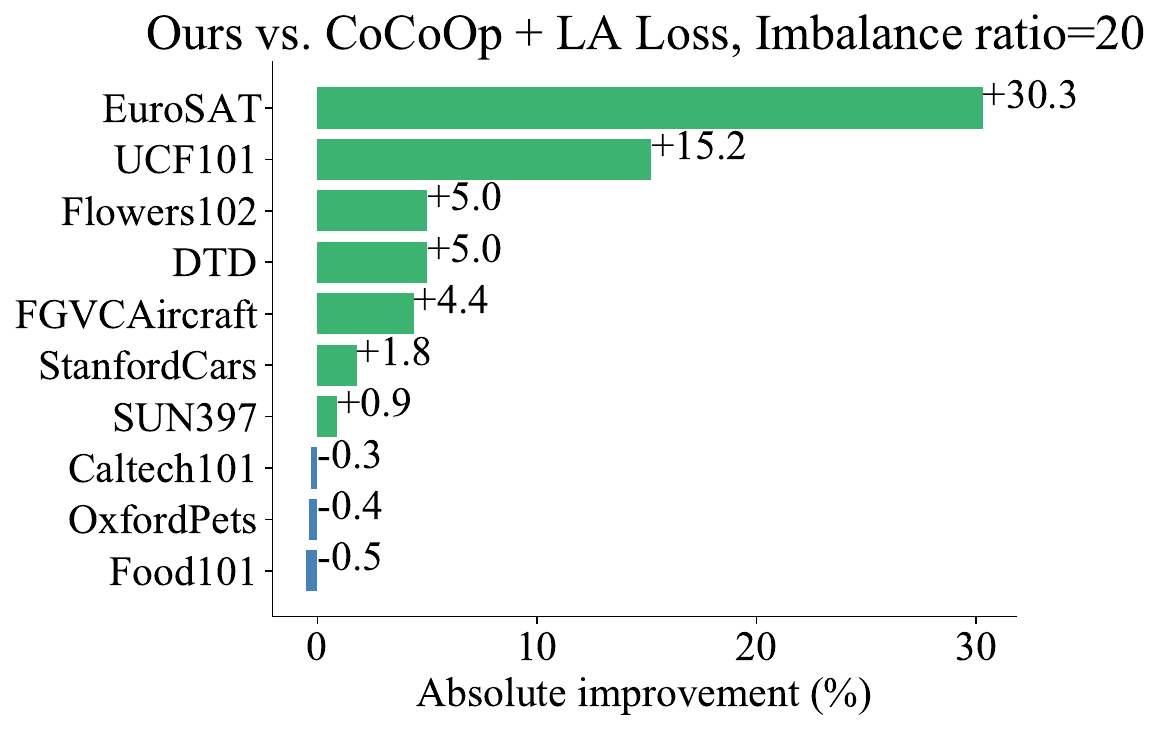}
    \includegraphics[width=0.31\textwidth]{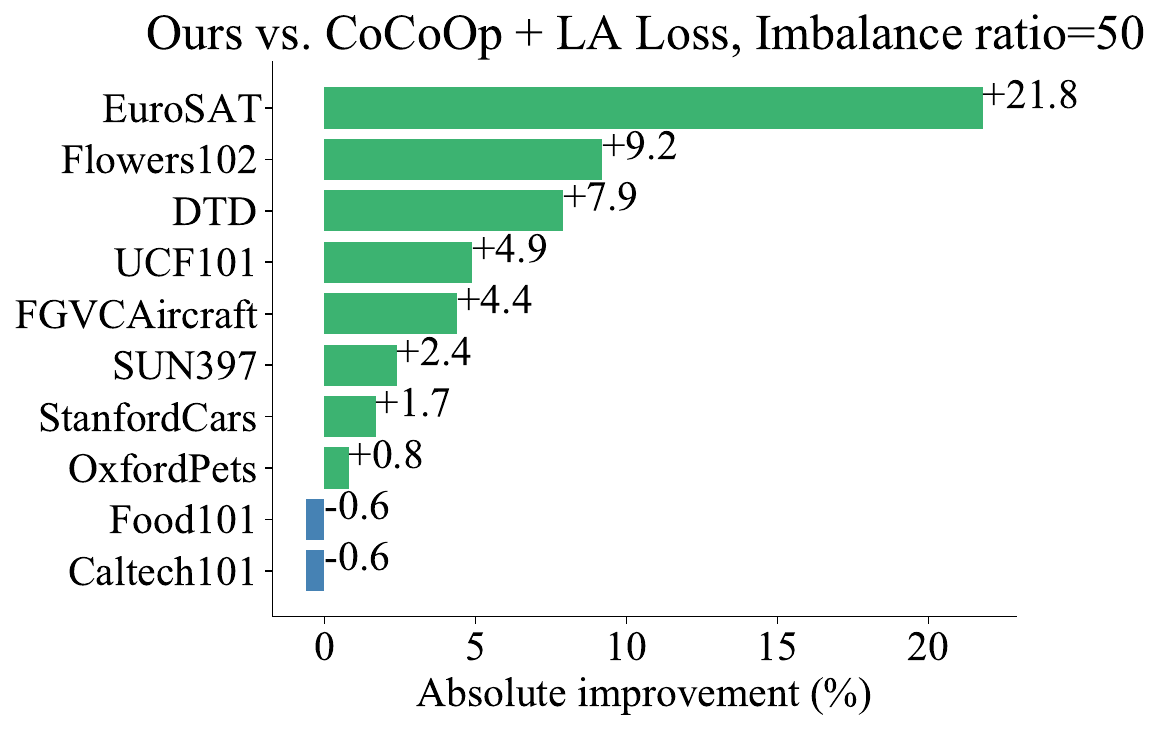}
    \vskip -0.1in
    \caption{Absolute improvement on the new classes with imbalance ratio 10, 20, 50}
    \label{fig:ib2n2}
\end{figure*}

\begin{table*}[!t]
\captionsetup{skip=0pt}
\caption{Comparison of different methods in 16-shot base-to-new generalization. We report the accuracy (\%)  on both base and new classes, as well as their harmonic mean. The best results are presented in bold.}
\vskip -0.05in
\label{tab:b2n}
\begin{subtable}{0.33\textwidth}
\caption{\textbf{Average over 11 datasets.}}
\begin{center}
\begin{tabular}{ccc|c}
\toprule
& Base & New & H \\ 
\midrule
CLIP   & 69.34 & 74.22 & 71.70 \\
CoOp   & 82.69 & 63.22 & 71.66 \\
CoCoOp & 80.47 & 71.69 & 75.83 \\
LFA    & 83.62 & 74.56 & 78.83 \\
\rowcolor[HTML]{E8E5E5} Ours & \textbf{83.86} & \textbf{76.55} & \textbf{80.04} \\
\bottomrule
\end{tabular}
\end{center}
\end{subtable}
\begin{subtable}{0.33\textwidth}
\caption{ImageNet.}
\begin{center}
\begin{tabular}{ccc|c}
\toprule
& Base & New & H \\ 
\midrule
CLIP   & 72.43 & 68.14 & 70.22 \\
CoOp   & 76.47 & 67.88 & 71.92 \\
CoCoOp & 75.98 & \textbf{70.43} & \textbf{73.10} \\
LFA    & 76.89 & 69.36 & 72.93 \\
\rowcolor[HTML]{E8E5E5} Ours & \textbf{76.97} & 68.54 & 72.48 \\ \bottomrule
\end{tabular}
\end{center}
\end{subtable}
\begin{subtable}{0.33\textwidth}
\caption{Caltech101.}
\begin{center}
\begin{tabular}{ccc|c}
\toprule
& Base & New & H \\ 
\midrule
CLIP   & 96.84 & 94.00 & 95.40 \\
CoOp   & 98.00 & 89.91 & 93.73 \\
CoCoOp & 97.96 & 93.81 & 95.84 \\
LFA    & 98.41 & 93.93 & 96.13 \\
\rowcolor[HTML]{E8E5E5} Ours & \textbf{98.54} & \textbf{94.47} & \textbf{96.46} \\ \bottomrule
\end{tabular}
\end{center}
\end{subtable}
\begin{subtable}{0.33\textwidth}
\caption{OxfordPets.}
\begin{center}
\begin{tabular}{ccc|c}
\toprule
& Base & New & H              \\ 
\midrule
CLIP   & 91.17 & 97.26 & 94.12 \\
CoOp   & 93.67 & 95.29 & 94.47 \\
CoCoOp & 95.20 & \textbf{97.69} & \textbf{96.43} \\
LFA    & 95.13 & 96.23 & 95.68 \\
\rowcolor[HTML]{E8E5E5} Ours & \textbf{95.53} & 97.34 & \textbf{96.43} \\
\bottomrule
\end{tabular}
\end{center}
\end{subtable}
\begin{subtable}{0.33\textwidth}
\caption{StanfordCars.}
\begin{center}
\begin{tabular}{ccc|c}
\toprule
& Base & New & H \\ 
\midrule
CLIP   & 63.37 & 74.89 & 68.85 \\
CoOp   & 78.12 & 60.40 & 68.13 \\
CoCoOp & 70.49 & 73.59 & 72.01 \\
LFA    & 76.32 & 74.88 & 75.59 \\
\rowcolor[HTML]{E8E5E5} Ours & \textbf{79.14} & \textbf{74.92} & \textbf{76.97} \\
\bottomrule
\end{tabular}
\end{center}
\end{subtable}
\begin{subtable}{0.33\textwidth}
\caption{Flowers102.}
\begin{center}
\begin{tabular}{ccc|c}
\toprule
& Base & New & H \\ 
\midrule
CLIP   & 72.08 & \textbf{77.80} & 74.83 \\
CoOp   & 97.60 & 59.67 & 74.06 \\
CoCoOp & 94.87 & 71.75 & 81.71 \\
LFA    & 97.34 & 75.44 & 85.00 \\
\rowcolor[HTML]{E8E5E5} Ours & \textbf{98.01} & 77.52 & \textbf{86.57} \\
\bottomrule
\end{tabular}
\end{center}
\end{subtable}
\begin{subtable}{0.33\textwidth}
\caption{Food101.}
\begin{center}
\begin{tabular}{ccc|c}
\toprule
& Base & New & H \\ 
\midrule
CLIP   & 90.10 & 91.22 & 90.66 \\
CoOp   & 88.33 & 82.26 & 85.19 \\
CoCoOp & \textbf{90.70} & 91.29 & 90.99 \\
LFA    & 90.52 & \textbf{91.48} & \textbf{91.00} \\
\rowcolor[HTML]{E8E5E5} Ours & 90.52 & 91.23 & 90.87 \\
\bottomrule
\end{tabular}
\end{center}
\end{subtable}
\begin{subtable}{0.33\textwidth}
\caption{FGVCAircraft.}
\begin{center}
\begin{tabular}{ccc|c}
\toprule
& Base & New & H \\ 
\midrule
CLIP   & 27.19 & 36.29 & 31.09 \\
CoOp   & 40.44 & 22.30 & 28.75 \\
CoCoOp & 33.41 & 23.71 & 27.74 \\
LFA    & 41.48 & 32.29 & 36.31 \\
\rowcolor[HTML]{E8E5E5} Ours & \textbf{43.86} & \textbf{36.69} & \textbf{39.96} \\
\bottomrule
\end{tabular}
\end{center}
\end{subtable}
\begin{subtable}{0.33\textwidth}
\caption{SUN397.}
\begin{center}
\begin{tabular}{ccc|c}
\toprule
& Base & New & H \\ 
\midrule
CLIP   & 69.36 & 75.35 & 72.23 \\
CoOp   & 80.60 & 65.89 & 72.51 \\
CoCoOp & 79.74 & 76.86 & 78.27 \\
LFA    & \textbf{82.13} & 77.20 & 79.59 \\
\rowcolor[HTML]{E8E5E5} Ours & 81.64 & \textbf{77.93} & \textbf{79.74} \\
\bottomrule
\end{tabular}
\end{center}
\end{subtable}
\begin{subtable}{0.33\textwidth}
\caption{DTD.}
\begin{center}
\begin{tabular}{ccc|c}
\toprule
& Base & New & H \\ 
\midrule
CLIP   & 53.24 & 59.90 & 56.37 \\
CoOp   & 79.44 & 41.18 & 54.24 \\
CoCoOp & 77.01 & 56.00 & 64.85 \\
LFA    & 81.29 & 60.63 & 69.46 \\
\rowcolor[HTML]{E8E5E5} Ours & \textbf{81.40} & \textbf{61.35} & \textbf{69.97} \\
\bottomrule
\end{tabular}
\end{center}
\end{subtable}
\begin{subtable}{0.33\textwidth}
\caption{EuroSAT.}
\begin{center}
\begin{tabular}{ccc|c}
\toprule
& Base & New & H \\ 
\midrule
CLIP   & 56.48 & 64.05 & 60.03 \\
CoOp   & 92.19 & 54.74 & 68.90 \\
CoCoOp & 87.49 & 60.04 & 71.21 \\
LFA    & \textbf{93.40} & 71.24 & 80.83 \\
\rowcolor[HTML]{E8E5E5} Ours & 89.97 & \textbf{81.33} & \textbf{85.43} \\ \bottomrule
\end{tabular}
\end{center}
\end{subtable}
\begin{subtable}{0.33\textwidth}
\caption{UCF101.}
\begin{center}
\begin{tabular}{ccc|c}
\toprule
& Base & New & H \\ 
\midrule
CLIP   & 70.53 & 77.50 & 73.85 \\
CoOp   & 84.69 & 56.05 & 67.46 \\
CoCoOp & 82.33 & 73.45 & 77.64 \\
LFA    & 86.97 & 77.48 & 81.95 \\
\rowcolor[HTML]{E8E5E5} Ours & \textbf{87.13} & \textbf{80.51} & \textbf{83.69} \\
\bottomrule
\end{tabular}
\end{center}
\end{subtable}
\end{table*}

\begin{table*}[!t]
\caption{Cross-dataset transfer learning accuracy (\%)  of different methods. The methods are trained on an imbalanced source dataset (ImageNet) and subsequently evaluated on the target
datasets. The best results are presented in bold.}
\vskip -0.05in
\label{tab:xd}
\begin{subtable}{\textwidth}
\begin{center}
\begin{tabular}{c|ccccccccccc} 
\toprule
& Cal. & OP. & SC. & FLw. & Food. & FA. & SUN. & DTD. & ES. & UCF. & Avg. \\ \midrule
CoOp + LogitAdjusted Loss   &  90.8    &  87.0   &  64.9   &  67.3    &  85.3     & 18.8   &  63.2    & 42.2     & 44.4    &  65.9    &  63.0    \\
CoCoOp + LogitAdjusted Loss &  \textbf{91.4}    &  88.6   &  \textbf{65.6}   &  \textbf{69.4}    &  \textbf{86.3}    &  23.0   &  66.0    &  \textbf{45.0}   &  42.8   &  \textbf{67.5}    &  64.6    \\
\rowcolor[HTML]{E8E5E5} 
\algo~(Ours)                          &  91.3   &  \textbf{88.9}   &  64.6   &  68.3    & 85.5    & \textbf{24.2}  & \textbf{66.1}    &  44.6  & \textbf{48.4}   &  67.2   &  \textbf{64.9}  \\ \bottomrule
\end{tabular}
\end{center}
\end{subtable}
\end{table*}

\textbf{Implementation Details.} 
We use ViT-B/16 as the vision backbone for all methods for fair comparison. Our models are trained for 10-100 epochs on each dataset and use the SGD optimizer with a batch size of 128, learning rate of $3 \times 10 ^{-4}$, weight decay of $5 \times 10 ^{-4}$, and momentum of 0.9. The temperature parameter $\tau_{t}$ for image-text matching is set to 0.01 following CLIP, whereas the $\tau_{v}$ for image-image matching is decided by searching from \{0.005, 0.01, 0.02, 0.05, 0.1\} on each dataset. For the baseline methods, the results are generated by following the exact setting as introduced in the original articles. All the experiments are carried out on a single NVIDIA GeForce RTX 3090.

\subsection{Main Results}

\textbf{Generalization from base to new classes.}
For generalization from base to new classes, we partition each dataset into two disjoint subsets, namely base classes and new classes. Then, the model is trained on the imbalanced base set and subsequently tested on base and new classes to demonstrate its generalization ability.  Table~\ref{tab:ib2n} presents the harmonic mean values for the base-to-new setting over imbalance ratios \{10, 20, 50\}. ImageNet is skipped here due to the extremely high training cost for CoCoOp under this setting. The results show that \algo\ consistently achieves state-of-the-art results across different imbalance ratios. Specifically, the harmonic mean values of \algo\ outperform the best previous method by an average of 3.67\%, 4.60\%, and 3.88\% under the three imbalance ratios. We also illustrate the absolute improvements of \algo\ compared to the previous best method in Figure~\ref{fig:ib2n1} and Figure~\ref{fig:ib2n2}. The results show average improvements of 2.58\%, 2.79\%, and 2.27\% on the base classes and higher average improvements of 4.87\%, 6.11\%, and 5.19\% on the new classes, affirming that it does indeed compensate for new classes. Together, the above results demonstrate the effectiveness of our approach in addressing imbalance both within the base classes and between the base and new classes. We present detailed results for each setting in the appendix due to the page limit.

To examine the robustness of \algo, we also report the results for 16-shot base-to-new generalization in Table~\ref{tab:b2n}. In this setting, \algo\ still achieves an improvement of 1.21\% in average harmonic mean over the best previous method. Specifically, it performs comparably with LFA on the base classes (+0.14\% by average) but outperforms LFA on the new classes by a large margin (+1.99\% by average), thus validating its ability to help with new classes. 

In addition, by taking a closer look at the results for each dataset, \algo\ achieves significant gains on datasets such as Flowers102, FGVCAircraft, EuroSAT, and UCF101. This is in accordance with the analysis in Section~\ref{sec: cma} that CLIP performs poorly on datasets where textual information is relatively unreliable, and our proposed approaches alleviate this issue by leveraging both visual and textual information.

\textbf{Cross dataset transfer.}
For cross-dataset transfer, we train the model on an imbalanced ImageNet subset with an imbalance ratio of 100 and subsequently test the model on the other 10 datasets. Table~\ref{tab:xd} presents the results for cross-dataset transfer. \algo\ shows similar results compared to CoCoOp with LogitAdjusted Loss across the 10 target datasets, achieving an average improvement of 0.3\%. It's worth noting that the baseline methods require much more training time compared to ours. For CoCoOp, 10 epochs of training lasts for 1 day and 6 hours and inference alone takes up 3 hours, while our method only needs about 20 minutes for the whole training process. Nonetheless, our method \algo\ is able to deliver comparable results while significantly reducing computational cost. Similar to the base-to-new generalization task, performance gains on specific datasets can be observed in this task as well.

\begin{table}[!t]
\caption{Domain generalization accuracy (\%)  of different methods. The methods are trained on an imbalanced source dataset (ImageNet) and subsequently evaluated on the target datasets. The best results are presented in bold.}
\vskip -0.05in
\label{tab:dg}
\begin{center}
\begin{tabular}{cc|cccc}
\toprule
& IN.  & IN-A. & INV2. & IN-S & IN-R. \\ \midrule
CoOp + LA Loss   & 70.7 &  48.7   &  \textbf{63.5}   &   47.2   &  73.8     \\
CoCoOp + LA Loss & 71.3 &  \textbf{49.1}    & 63.3   &  47.8  &   74.4    \\ \midrule
\rowcolor[HTML]{E8E5E5} 
\algo~(Ours)           & \textbf{71.6} &  \textbf{49.1}    &  62.8     &  \textbf{48.3}    &  \textbf{75.0}    \\ \bottomrule
\end{tabular}
\end{center}
\end{table}

\textbf{Domain generalization.}
For domain generalization, we train the model on an imbalanced ImageNet subset with an imbalance ratio of 100 and evaluate the model on four domain-shift target datasets. The results are presented in Table~\ref{tab:dg}. \algo\ achieves improvements over the previous best method on 3 out of 4 target datasets, with an average increase of 0.15\% on the target datasets, and an increase of 0.3\% on the source dataset. The results demonstrate the robustness of \algo\ against domain shifts.

\subsection{Ablation Study}

\textbf{Impacts of different loss functions.}
In the main results, we equip the baseline methods with the balanced LA loss for fair comparison. Here, we further examine the robustness of different methods against class imbalance without the assistance of such a tailored loss function. Specifically, we use cross entropy (CE) loss for all the methods and run on the imbalanced base-to-new generalization task. The results are shown in Table~\ref{tab:ablation}. It can be observed that CoCoOp suffers from a more severe performance drop without LA loss, i.e., a decrease of 5.78\% in average harmonic value. In contrast, our method \algo\ manages to hold on with a drop of only 1.51\%. This indicates that even without the balanced logit-adjusted loss, our model still shows potential strength in dealing with class imbalance.

\begin{table}[!t]
\caption{Ablation on different loss functions. $\Delta$ indicates the difference in performance for the same method trained with CE loss or LA loss. Our method is the least sensitive to the change in loss function.}
\vskip -0.05in
\label{tab:ablation}
\begin{center}
\begin{tabular}{cccc}
\toprule
& Base & New & Harmonic Mean \\ \midrule
\multicolumn{1}{c|}{CoOp + LA Loss}                     &  80.26    & 61.94    &   69.92           \\
\multicolumn{1}{c|}{CoOp + CE Loss}                     &  76.12    &  61.19   &    67.84           \\ \midrule
\rowcolor[HTML]{E8E5E5} 
\multicolumn{1}{c|}{\cellcolor[HTML]{E8E5E5}$\Delta$} &  -4.14\%    & -0.75\%    &   -2.08\%            \\ \midrule
\multicolumn{1}{c|}{CoCoOp + LA Loss}                   &  77.91    &  71.05   &     74.32          \\
\multicolumn{1}{c|}{CoCoOp + CE Loss}                   &  72.05    &  65.36   &     68.54          \\ \midrule
\rowcolor[HTML]{E8E5E5} 
\multicolumn{1}{c|}{\cellcolor[HTML]{E8E5E5}$\Delta$} &  -5.86\%    &  -5.69\%   &   -5.78\%            \\ \midrule
\multicolumn{1}{c|}{\algo~(Ours)}                     &  80.38    &  76.14   &  78.20             \\
\multicolumn{1}{c|}{\algo\ w/ CE Loss}                    &  78.07    &  75.36   &  76.69             \\ \midrule
\rowcolor[HTML]{E8E5E5} 
\multicolumn{1}{c|}{\cellcolor[HTML]{E8E5E5}$\Delta$} &  -2.31\%    & -0.78\%    &  -1.51\%             \\ \bottomrule
\end{tabular}
\end{center}
\end{table}

\begin{figure}[!t]
\centering
\includegraphics[width=0.49\linewidth]{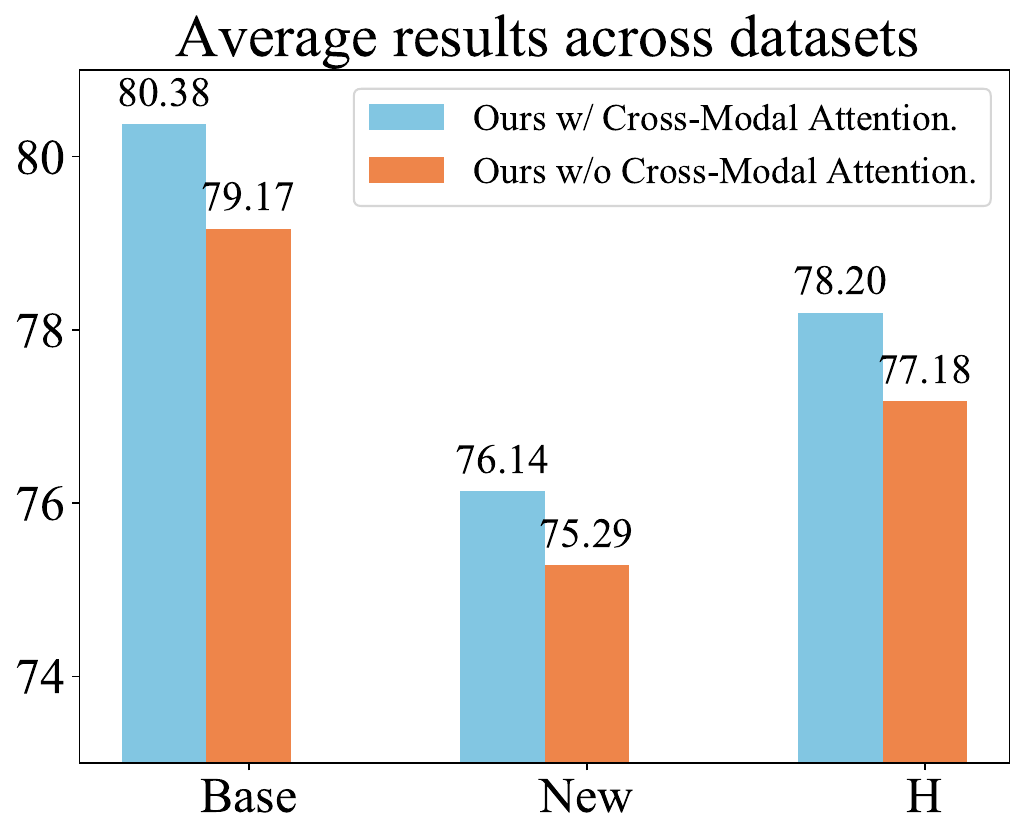}
\includegraphics[width=0.49\linewidth]{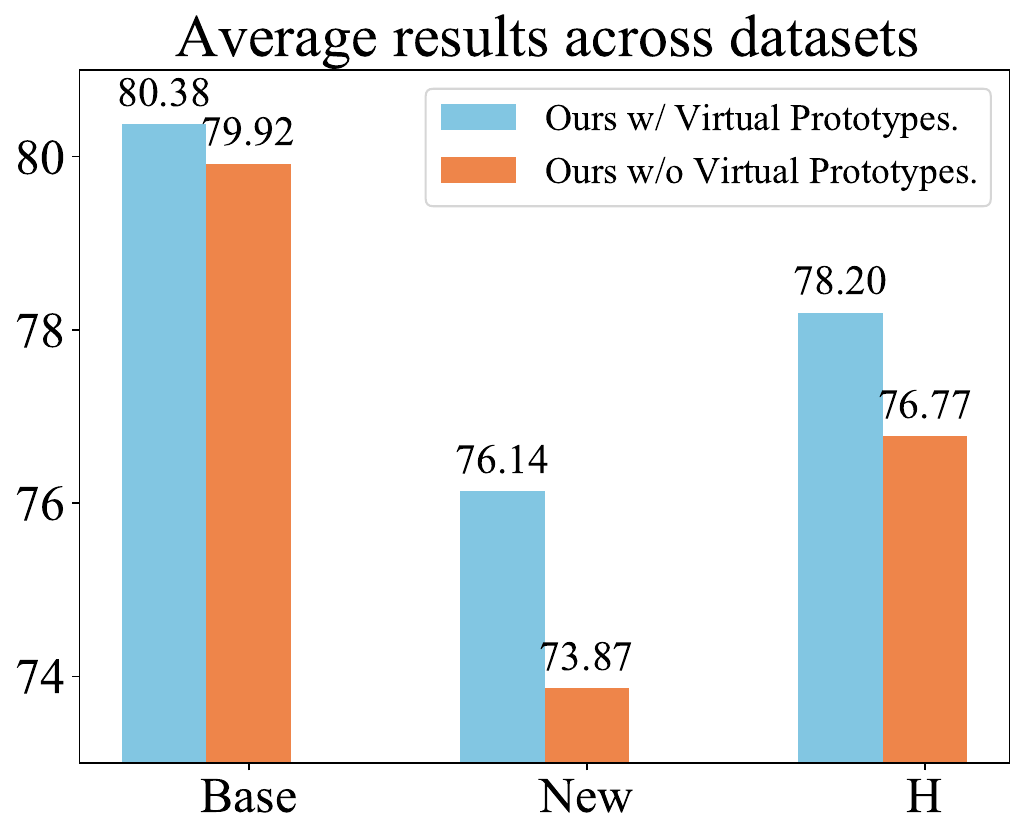}
\vskip -0.05in
\caption{Ablation studies on cross-modal attention (left) and virtual prototypes (right). The experiment is conducted on the imbalanced base-to-new generalization task with an imbalance ratio of 50.}
\label{fig:ablation}
\end{figure}

\textbf{Effect of cross-modal attention.}
We conduct ablation study on the imbalanced base-to-new generalization task to examine the effectiveness of the cross-modal attention module. For the sake of simplicity, we examine on the imbalance ratio = 50 setting. The current model is compared to one with only linear projection after the extracted features, with the rest of the settings the same. The results are shown in the left part of Figure~\ref{fig:ablation}. Without the cross-modal attention module, the average results on the base and new classes experience a drop of 1.19\% and 0.85\% respectively, leading to a 1.02\% decline of harmonic mean value. These figures clearly show that our cross-modal attention module acts contributes positively and significantly to the model's overall performance.

\textbf{Effect of virtual prototypes.}
We further examine the effectiveness of the virtual prototypes. Since the removal of virtual prototypes renders the image-image matching for new classes unachievable, the model in comparison can only leverage image-text matching on the new classes. We conduct comparison experiments and report the results in the right part of Figure~\ref{fig:ablation}. The results show that, the performance gap on the base classes is relatively small, with our proposed model holding an average advantage of 0.46\%. However, the performance on the new classes drops remarkably in response to the removal of virtual prototypes, showing an average decline of 2.27\% across different datasets. The results prove that the introduction of virtual prototypes significantly helps with new class generalization.

\section{Conclusion}
In this paper, we aim to address the new class generalization problem for vision-language models under more practical scenarios, where the data may exhibit a long-tailed distribution. We propose a novel and simple framework named \algo\ to solve this issue in an efficient manner. \algo\ achieves state-of-the-art performance over extensive experiments on diverse image classification datasets, with an especially strong generalization on the new classes. Just as significantly, the proposed framework directly optimizes in the feature space and does not need access to model weights, which also contributes to its economical training cost compared to past methods. We hope our work serves as an inspiration for further advances in exploring efficient and long-tailed generalization for vision-language models.

\begin{acks}
This research was supported by the National Science Foundation of China (62176118).
\end{acks}

\bibliographystyle{ACM-Reference-Format}
\bibliography{reference}

\appendix

\begin{table*}[!t]
\captionsetup{skip=0pt}
\caption{Base-to-new generalization results of different methods with imbalance ratios 10, 20, 50. The best results are in bold.}
\label{tab:ib2n-add}
\begin{subtable}{\textwidth}
\caption{Imbalance Ratio = 10, base classes} 
\begin{center}
\begin{tabular}{c|ccccccccccc} 
\toprule
& Cal. & OP. & SC. & FLw. & Food. & FA. & SUN. & DTD. & ES. & UCF. & Avg. \\
\midrule
CoOp + LogitAdjusted Loss   &  96.5    &  94.7   &  \textbf{75.8}   &  \textbf{98.7}    &  89.4     &  39.0  &   74.5   &  \textbf{81.0}    &  \textbf{94.3}   &  83.9   &  82.8    \\
CoCoOp + LogitAdjusted Loss &  94.5    &  \textbf{95.6}   &  70.2   &   94.6   & \textbf{90.7}      &  34.9   &  78.7   &  75.2    &  87.8   &  81.6    &  80.5    \\
\rowcolor[HTML]{E8E5E5} 
\algo~(Ours)                          &  \textbf{96.9}   &  95.0   &  74.2   &  98.0     &   90.5    &  \textbf{40.0}   &  \textbf{81.1}    & 80.8     &  89.0    &   \textbf{87.2}    &  \textbf{83.3}     \\
\bottomrule
\end{tabular}
\end{center}
\end{subtable} 
\begin{subtable}{\textwidth}
\caption{Imbalance Ratio = 10, new classes}
\begin{center}
\begin{tabular}{c|ccccccccccc}
\toprule
& Cal. & OP. & SC. & FLw. & Food. & FA. & SUN. & DTD. & ES. & UCF. & Avg. \\ \midrule
CoOp + LogitAdjusted Loss   &  87.5   &  93.5  &  63.7   &  56.5    &   89.1    &  27.3   &  59.1   &  41.5    & 41.2    & 57.0    &  61.6    \\
CoCoOp + LogitAdjusted Loss & 94.3    &  \textbf{97.8}  &  73.7   &  65.8  & \textbf{91.7}     &  32.6   &  77.3   &   57.4   &  45.9   &  72.5     &  70.9    \\
\rowcolor[HTML]{E8E5E5} 
\algo~(Ours)                          & \textbf{94.9}    &  97.0  &  \textbf{74.4}   & \textbf{75.1}      &  91.1     &  \textbf{35.8}   &  \textbf{77.5}    &  \textbf{58.9}    &  \textbf{73.5}    & \textbf{79.5} & \textbf{75.8}  \\ \bottomrule
\end{tabular}
\end{center}
\end{subtable} 
\begin{subtable}{\textwidth}
\caption{Imbalance Ratio = 20, base classes}
\begin{center}
\begin{tabular}{c|ccccccccccc}
\toprule
& Cal. & OP. & SC. & FLw. & Food. & FA. & SUN. & DTD. & ES. & UCF. & Avg. \\
\midrule
CoOp + LogitAdjusted Loss & \textbf{96.5} & \textbf{93.9} & 73.1 & \textbf{98.5}    &   88.9    &  37.0   &  77.6   &  77.0    & \textbf{93.1}    &  83.7    &  81.9    \\
CoCoOp + LogitAdjusted Loss &  95.2    &  \textbf{95.9}  &  69.2   &  94.0    & \textbf{90.8}     &  33.6   &  78.2    &   71.6   &  88.3   &  80.7    &  79.8   \\
\rowcolor[HTML]{E8E5E5} 
\algo~(Ours)                          & \textbf{96.7}    &  94.8  &  72.5   & 97.7     &  90.5     &  \textbf{39.8}   &  \textbf{79.8}    &  \textbf{79.1}    &  90.5    & \textbf{85.7} & \textbf{82.5}  \\
\bottomrule
\end{tabular}
\end{center}
\end{subtable}
\begin{subtable}{\textwidth}
\caption{Imbalance Ratio = 20, new classes}
\begin{center}
\begin{tabular}{c|ccccccccccc}
\toprule
& Cal. & OP. & SC. & FLw. & Food. & FA. & SUN. & DTD. & ES. & UCF. & Avg. \\ \midrule
CoOp + LogitAdjusted Loss   &  89.1   &  94.4  &  62.5   &  58.9    &   84.0    &  24.3   &  62.0   &  43.0    & 47.2    &  46.9    &  61.2    \\
CoCoOp + LogitAdjusted Loss & \textbf{95.3}    &  \textbf{97.4}  &  73.7   &  70.1   & \textbf{91.6}     &  32.0   &  76.4   &   53.6   &  44.1   &  64.5    &  69.9    \\
\rowcolor[HTML]{E8E5E5} 
\algo~(Ours)                          & 95.0    &  97.0  &  \textbf{74.5}   & \textbf{75.1}      &  91.0     &  \textbf{36.4}   &  \textbf{77.3}    &  \textbf{58.6}    &  \textbf{74.4}    & \textbf{79.7} & \textbf{75.9}  \\ \bottomrule
\end{tabular}
\end{center}
\end{subtable}
\begin{subtable}{\textwidth}
\caption{Imbalance Ratio = 50, base classes}
\begin{center}
\begin{tabular}{c|ccccccccccc}
\toprule
& Cal. & OP. & SC. & FLw. & Food. & FA. & SUN. & DTD. & ES. & UCF. & Avg. \\ \midrule
CoOp + LogitAdjusted Loss   &  93.8    &  92.3  &  \textbf{70.7}   &  \textbf{96.6}    &   88.0    &  35.3   &  \textbf{79.7}   &  \textbf{72.0}    & \textbf{93.4}    &  80.8    &  80.3    \\
CoCoOp + LogitAdjusted Loss &  94.5    &  \textbf{94.6}  &  67.1   &  88.5    & \textbf{90.5}     &  31.5   &  77.3    &   68.3   &  86.9   &  79.9   &  77.9    \\
\rowcolor[HTML]{E8E5E5} 
\algo~(Ours)                          & \textbf{95.2}    &  \textbf{94.6}  &  69.0   & \textbf{95.3}      &  90.3     &  \textbf{37.6}   &  78.6   &  \textbf{72.0}    &  87.9    & \textbf{83.3} & \textbf{80.4}  \\ \bottomrule
\end{tabular}
\end{center}
\end{subtable}
\begin{subtable}{\textwidth}
\caption{Imbalance Ratio = 50, new classes}
\begin{center}
\begin{tabular}{c|ccccccccccc}
\toprule
& Cal. & OP. & SC. & FLw. & Food. & FA. & SUN. & DTD. & ES. & UCF. & Avg. \\
\midrule
CoOp + LogitAdjusted Loss   &  92.8    &  94.4  &  64.0   &  61.9    &   87.3   &  24.9   &  65.9   &  40.0    & 38.4    &  49.8    &  61.9    \\
CoCoOp + LogitAdjusted Loss &  \textbf{95.3}    &  96.3  &  73.1   &  67.9    & \textbf{91.7}     &  31.4   &  75.1    &   52.5   &  51.9   &  75.3    &  71.0  \\
\rowcolor[HTML]{E8E5E5} 
\algo~(Ours)                          & 94.7   &  \textbf{97.1}  &  \textbf{74.8}   & \textbf{76.1}      &  91.1     &  \textbf{35.8}   &  \textbf{77.5}    &  \textbf{60.4}    &  \textbf{73.7}    & \textbf{80.2} & \textbf{76.1}  \\
\bottomrule
\end{tabular}
\end{center}
\end{subtable}
\end{table*}

\newpage

\section{Additional Results}

\textbf{Generating imbalanced datasets.}
Following the method proposed by \citet{cao2019learning}, we generate imbalanced versions from the original datasets to obey an exponential decay of a given ratio. Let $n_{i}$ denote the number of samples in the $i$-th class, the imbalance ratio is defined as $\max\{n_{i}\} / \min \{n_{i}\}$. From Table~\ref{tab:data1}, we can see that the training set for some datasets only has around 30 samples per class (FGVCAircraft) whereas some has over 1000 (ImageNet). Therefore, we set the maximum number of samples per class of the generated dataset to be either 100 (if has) or the maximum number of samples per class of the original dataset, to guarantee enough data to form a valid imbalanced distribution. Additionally, in cases where the maximum number of samples per class is lower than the imbalance ratio, we ensure there is at least 1 sample instead of 0 for the tail classes.

\textbf{Imbalanced base-to-new generalization details.}
In Table~\ref{tab:ib2n-add}, we show the full results of imbalanced base-to-new generalization, including the accuracy of different methods on base and new classes, under imbalance ratios $\{10, 20, 50\}$. On the base classes, our method exhibits a slight edge over CoOp+LA Loss with an increase of $0.3\%$ averaging across different ratios and shows a clearer improvement ($2.7\%$) over CoCoOp+LA Loss, which is the previous best method in harmonic mean value. On the new classes, our method far outperforms CoOp+LA Loss with an advantage of $14.4\%$ and still leads CoCoOp+LA Loss by $5.3\%$. This again demonstrates that our method compensates significantly for the new classes while preserving a strong performance on the base classes.

\textbf{Different attention strategies.}
As mentioned in the article, we perform cross-modal attention by concatenating image features, visual prototypes and textual prototypes together, and then feed them into a self-attention module. Here, we provide analysis of different attention strategies by adding different input masks. For the sake of simplicity, we consider the concatenated inputs to be comprised of two parts, the visual part (image features + visual prototypes) and the texual part (texual prototypes). Hence, there are three different kinds of masks to choose from, including masking attention within each part and between each part. Table~\ref{tab:attn} shows the results of comparing different attention strategies with the original method (no mask at all). The consistent decline proves the superiority of the original design. Particularly, the removal of attention between visual and textual part leads to the largest drop on both base and new classes, which goes to show that the interaction between different modalities does contribute to improving the model's performance.

\begin{table}[!t]
\caption{Results of different attention strategies compared to the original method under the imbalanced base-to-new generalization setting with imbalance ratio $50$. $\Delta$ indicates the average difference in accuracy across different datasets. }
\label{tab:attn}
\begin{center}
\begin{tabular}{c|cc}
\toprule
Mask Type               & $\Delta$, Base Classes & $\Delta$, New Classes \\ \midrule
Within Visual.          &  -0.26\%    &  -0.49\%   \\
Within Text.            &  -0.19\%    &  -0.78\%   \\
Between Visual \& Text. &  -0.74\%   &   -0.90\%  \\ \bottomrule
\end{tabular}
\end{center}
\end{table}

\end{document}